\documentclass[lettersize,journal]{IEEEtran}
\usepackage{amsmath,amsfonts}
\usepackage{algorithmic}
\usepackage{algorithm}
\usepackage{array}
\usepackage{amsmath}
\usepackage{amssymb}
\usepackage{amsthm}
\usepackage{xurl}
\usepackage{hyperref}
\usepackage[caption=false,font=normalsize,labelfont=sf,textfont=sf]{subfig}
\usepackage{textcomp}
\usepackage{multirow,booktabs}
\usepackage{stfloats}
\usepackage{verbatim}
\usepackage{amsmath,amssymb,amsfonts}
\usepackage{algorithmic}
\usepackage{xcolor}

\usepackage{graphicx}
\newtheorem{prop}{Proposition} 
\begin{document}
\title{Beyond Similarity Entropy: A Rank-Guided Data-Free Quantization Framework for Vision Transformers}
\author{Runqing Jiang, Ye Zhang, Pengpeng Yu, Zhuo Song, Changjun Li, Longguang Wang, Yulan Guo,~\IEEEmembership{Senior Member,~IEEE}
\thanks{%
R. Jiang, P. Yu, Z. Song, Y. Zhang, C. Li, and Y. Guo
are with the School of Electronics and Communication Engineering,
Shenzhen Campus, Sun~Yat-sen~University, Shenzhen, China
(e-mail: \protect\nolinkurl{jiangrq3@mail2.sysu.edu.cn};
\protect\nolinkurl{yupp5@mail2.sysu.edu.cn};
\protect\nolinkurl{songzh28@mail2.sysu.edu.cn};
\protect\nolinkurl{zhangy2658@mail.sysu.edu.cn};
\protect\nolinkurl{lichj68@mail2.sysu.edu.cn};
\protect\nolinkurl{guoyulan@sysu.edu.cn}).%
}%

\thanks{
L. Wang is with the College of Electronic Science and Technology, Aviation University of Air Force, Changchun, China, 130012. (E-mail: wanglongguang15@nudt.edu.cn).
}

\thanks{Corresponding author: Ye Zhang.}
}

\maketitle
\begin{abstract}
Vision Transformers (ViTs) have demonstrated remarkable performance, yet their computational and memory demands pose significant challenges for deployment on mobile devices.
Data-free quantization (DFQ) enables the compression of ViTs without access to their original training data. 
Existing DFQ methods often synthesize calibration samples by maximizing the entropy of patch similarity. 
However, similarity entropy describes the dispersion of similarity values without ensuring that the generated samples cover diverse feature directions, potentially limiting their effectiveness for quantization calibration. 
To address this limitation, we propose RankQ-ViT, a rank-guided DFQ framework for ViTs. We first observe that patch similarity matrices derived from Gaussian noise tend to have lower ranks than those derived from real images. 
Motivated by this observation, we introduce a block-aware self-contrastive objective that promotes feature diversity during synthetic data generation without explicitly computing matrix rank at each optimization step.
We then select calibration samples based on their rank and semantic consistency and use them for block-wise reconstruction of the quantized model. 
Experiments on ImageNet, COCO, ModelNet40, and ShapeNet demonstrate the effectiveness of RankQ-ViT across multiple architectures and tasks, particularly under low-bit quantization. The code is available at \url{https://github.com/Runqing-forMost/rankq-vit.git}.

\end{abstract}


\section{Introduction}
\IEEEPARstart{V}{ision} Transformers (ViTs) have recently shown impressive performance in a wide range of computer vision tasks \cite{touvron2021training,xu2023improving,liu2021swin}. Despite their success, the cutting-edge ViT models are often characterized by intricate architectures and a substantial number of parameters, which severely hinders their deployment on resource-constrained edge devices. 
To improve the efficiency of ViTs, numerous model compression works have been developed, including compact architecture design \cite{liu2021swin,huang2020lightweight,xu2023devit}, network pruning \cite{li2025adaptive,xu2025lightweight,yi2025edgemoe}, knowledge distillation \cite{jiang2023knowledge,rao2023parameter,chen2025federated}, and model quantization \cite{chen2022energy,xie2025nestquant,cui2025lightweight,guo2025deep}. 
These works significantly speed up and reduce the storage requirements of ViT models, while maintaining stable performance. 

\begin{figure}[t] 
	\centering 
	\includegraphics[width=0.49\textwidth]{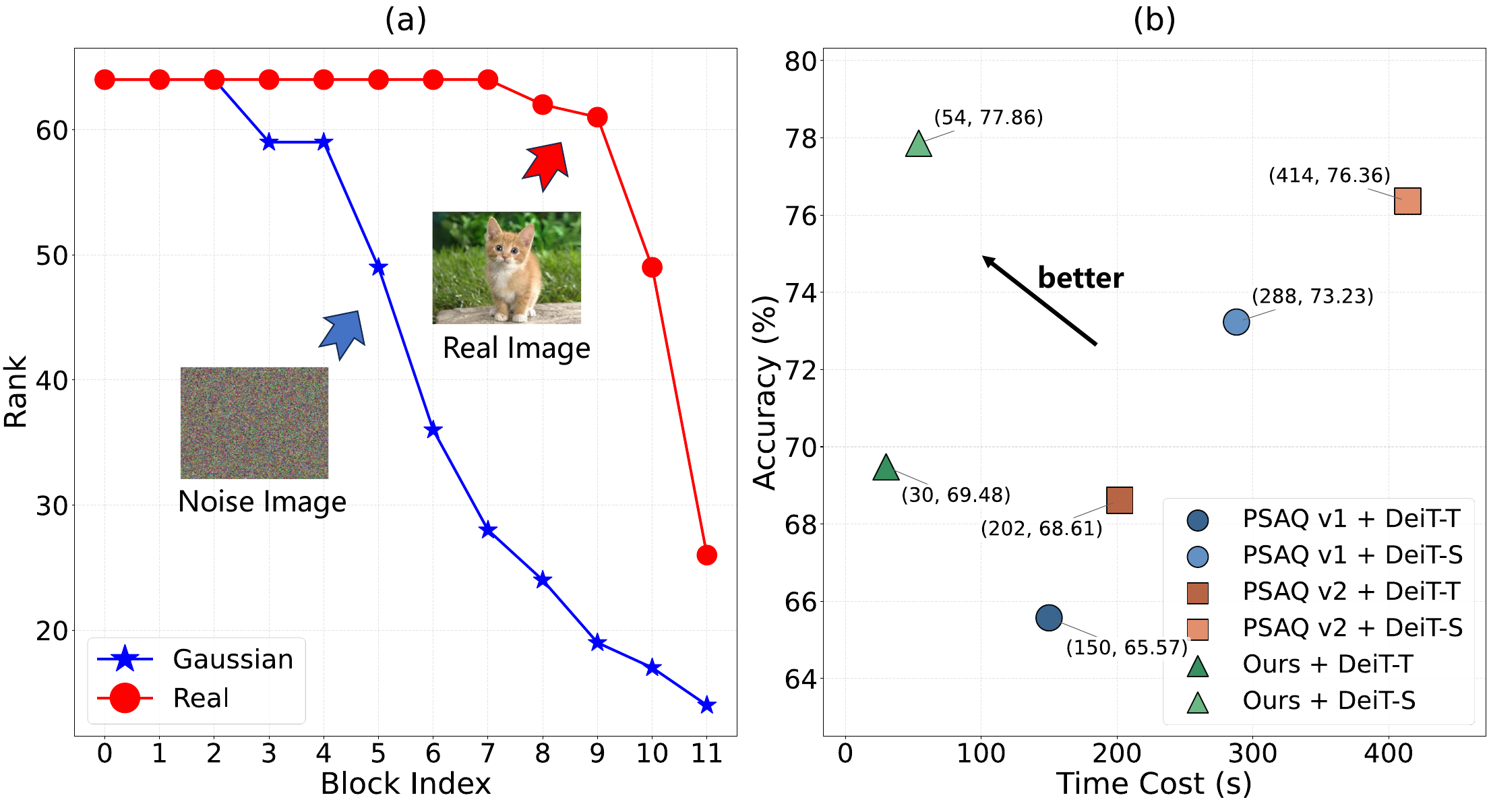} 
	\caption{Motivation and performance overview. (a) Rank distribution of the patch similarity matrix across different blocks, indicating that realistic images have higher ranks than noise ones. It motivates us to explicitly increase the ranks of the synthetic samples during sample generation.
    (b) Test accuracy (\%) and time cost of data generation of RankQ-ViT, PSAQ-ViT v1 \cite{li2022patch}, and PSAQ-ViT v2 \cite{li2023psaq} on DeiT-T and DeiT-S.} \label{fig:intro}
\end{figure}
Model quantization, which involves mapping floating-point numbers to fixed-point counterparts, has experienced growing interest due to its hardware-agnostic versatility.
Existing works \cite{li2022q,li2023repq,librecq,wang2022learnable} typically assume access to training datasets for calibrating quantized models, a paradigm known as quantization-aware training (QAT). While QAT achieves superior performance, it incurs massive training costs, especially for complex models like ViTs.
Some other works \cite{nagel2020up,zhong2025towards,li2026scalepredictor} adopt the post-training quantization (PTQ) paradigm, which only uses a small part of the training data, and learns the quantization parameters without the end-to-end retraining. 
However, the training datasets are not always available due to data privacy concerns \cite{li2022patch,zhang2026d4c}, rendering these methods inapplicable. To address this, a handful of recent studies \cite{cai2020zeroq,li2023hard,fan2024data} have begun to explore data-free quantization (DFQ), a paradigm that seeks to quantize pre-trained networks without relying on original training data.


The prevalent pipeline of DFQ involves generating synthetic samples from a pre-trained model, which are carefully crafted to mimic the distribution of the original training data, and then performing quantization with these synthetic samples. Some DFQ-based works \cite{cai2020zeroq,he2021generative} leverage statistics embedded in batch normalization (BN) layers as priors to guide synthetic sample generation. However, this strategy falters when applied to ViTs quantization, due to the absence of BN layers. To exploit the information encoded in pretrained ViTs, PSAQ-ViT and its successor \cite{li2022patch,li2023psaq} synthesize calibration samples by maximizing the entropy of patch similarity. However, the dispersion of similarity values does not ensure that the generated patch features cover diverse directions, potentially limiting their effectiveness for quantization calibration. Moreover, applying the same entropy objective across transformer blocks overlooks differences in their feature representations, while repeated probability density estimation increases the cost of data generation.

\par
To handle these issues, this paper proposes RankQ-ViT, an efficient and effective DFQ method for ViTs. Specifically, RankQ-ViT involves a two-step optimization framework. First, since the matrix rank of realistic images is consistently higher than that of noise ones (Fig.~\ref{fig:intro}(a)), we propose a block-aware rank promotion mechanism that explicitly raises the rank of the similarity matrix for generated images during optimization. This is achieved via a self-contrastive learning framework: patches from the same region (intra-patch) are pulled together, whereas patches from different regions (inter-patch) are pushed apart. This contrastive operation reduces linear correlations among patch features, increasing the matrix rank. We additionally provide a theoretical justification that this contrastive learning formulation effectively promotes the rank of the similarity matrix. Second, we propose a sample selection strategy on the generated samples to select reliable samples for the subsequent model quantization. We evaluate the quality of each generated sample using rank information and semantic consistency, and then select a portion of samples with highest quality score. These samples are then used for a block-wise reconstruction procedure, where the outputs of quantized ViTs are encouraged to mimic their full-precision counterparts in a block-wise manner, achieving both high training efficiency and better generalization performance, as indicated in Fig.~\ref{fig:intro}(b).

To summarize, our key contributions are listed as follows:
\begin{itemize}
    \item We first reveal that the rank of the similarity matrix serves as an effective indicator of image quality. Building on this insight, we propose RankQ-ViT, which incorporates a block-aware rank promotion mechanism along with a sample selection strategy to enable accurate quantization.

    \item  We develop a block-aware rank promotion mechanism based on self-contrastive learning to mitigate the distribution discrepancy between realistic images and generated samples.
    
    \item We propose a sample selection strategy, which utilizes the rank information and semantic consistency as the metric to select reliable samples for the model quantization.

    \item Experiments on four benchmark datasets (including ImageNet, COCO, ModelNet40, and ShapeNet) with various quantization regimes demonstrate the effectiveness and efficiency of our RankQ-ViT. Specifically, compared with the PSAQ-ViT, our method obtains over 10\% better results on ImageNet with W4/A4 quantization, while achieving around 5$\times$ faster data generation speed.
    
\end{itemize}

\begin{figure}[!t] 
	\centering 
	\includegraphics[width=0.49\textwidth]{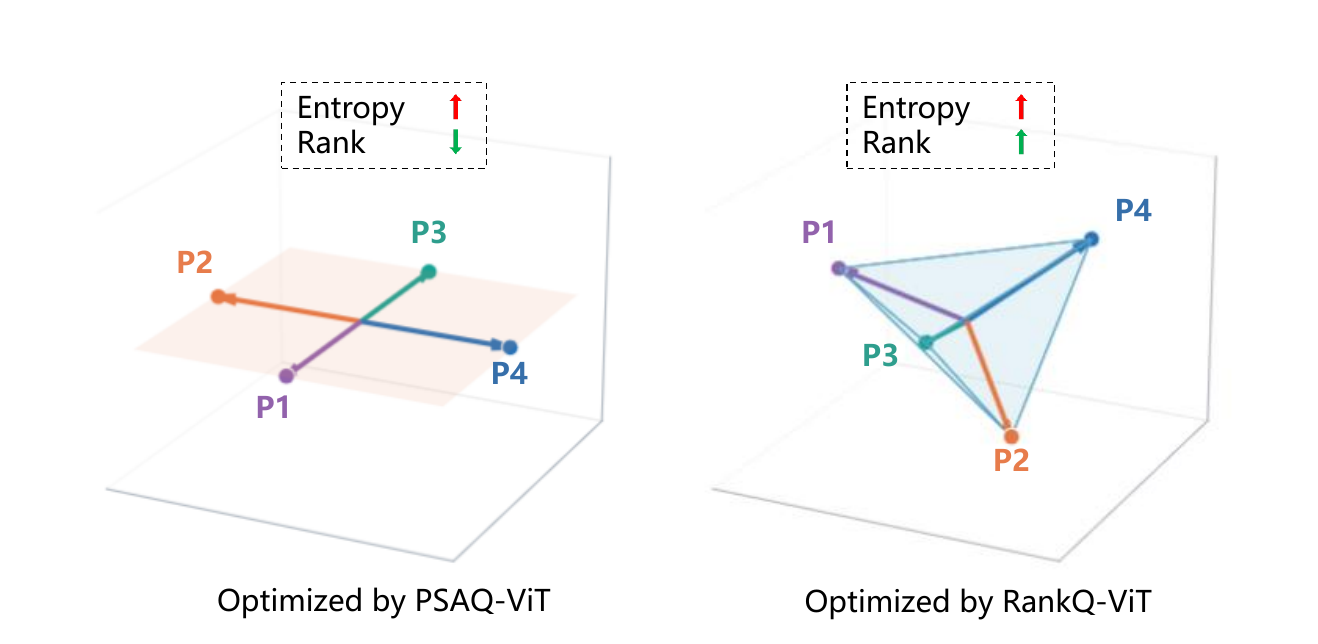} 
	\caption{Illustration of the distinction between patch-similarity entropy and feature-space rank. On the left, four normalized patch features lie in a two-dimensional plane and yield dispersed pairwise cosine similarities. On the right, four features span three dimensions despite having identical pairwise cosine similarities distributions. 
    This example shows that dispersion of similarity values alone does not guarantee coverage of independent feature directions, motivating rank-aware synthetic data generation for quantization calibration.} \label{fig:psaq_vs_rankq}
\end{figure}

\section{Related Work}

\subsection{Vision Transformers and Their Applications}

Dosovitskiy \textit{et al.} \cite{dosovitskiy2020image} designate the original ViT, which employs a pure transformer architecture and demonstrates superior performance to several off-the-shelf CNNs in image classification tasks. Later, Touvron \textit{et al.} \cite{touvron2021training} propose DeiT, which introduces a novel distillation strategy, wherein a well-trained CNN model serves as a teacher to guide the training of a transformer model. Such an approach enables the transformer to inherit valuable representations from the teacher model, achieving competitive performance with the top-tier CNN architectures. Based on this, Liu \textit{et al.}  \cite{liu2021swin} propose the Swin-Transformer. Swin-Transformer innovatively involves a window attention mechanism, which confines self-attention to a local window, significantly reducing computational costs and achieving superior performance. 

In addition to 2D images, several recent studies \cite{zhao2021point,guo2021pct} have demonstrated the great potential of transformer architecture in the realm of 3D point cloud processing. For example, Zhao \textit{et al.} \cite{zhao2021point} propose Point Cloud Transformer (PCT), which pioneeringly introduces the transformer architecture into 3D point cloud analysis, and achieves cutting-edge performance on point cloud classification and part segmentation tasks. Despite their impressive performance on various vision tasks, the intensive computation burden of these complicated ViTs significantly impedes their widespread applications on resource-constrained devices. In this paper, we are motivated to develop a quantization framework tailored for ViTs that operates without access to training data, thereby ensuring privacy and flexibility in real-world applications.

\subsection{Model Quantization}
In this section, we mainly recap the works on the post-training quantization and data-free quantization, which are closely related to our method.
\subsubsection{Post-training Quantization (PTQ)}
PTQ-based works quantize the neural networks using a small-scale calibration set, without heavily end-to-end retraining. Nagel \textit{et al.} \cite{nagel2020up} design AdaRound to learn the rounding function adaptively, with a layer-wise reconstruction paradigm. Li \textit{et al.} \cite{librecq} propose to perform reconstruction in a block-wise manner, attaining impressive image classification accuracy with 4-bit ResNet. In addition, some recent works have explored the PTQ for vision transformers. Liu \textit{et al.} \cite{liu2021post} first explore the challenges of post-training quantization for ViTs, and devise a ranking-aware distillation loss to preserve the ranking information after quantization. Li  \textit{et al.} \cite{li2023repq} propose RepQ-ViT, which uses scale reparameterization to decouple quantization from inference, applying complex quantizers for accurate calibration and transforming them into hardware-friendly simplified quantizers for efficient deployment. Jiang \textit{et al.} \cite{jiang2025aiqvit} use an architecture-informed mechanism to compensate for the quantization error of weights in ViTs, achieving superior performance on 2D and 3D recognition tasks. Despite their effectiveness, these methods still rely on a small amount of data for calibration. 

\subsubsection{Data-free Quantization}
DFQ-based works aim to quantize models without accessing the original training data, offering a more flexible way for model compression~\cite{li2026dvd,zhao2025enhancing,zhong2025semantic}. 
To achieve this, several methods \cite{li2023hard,cai2020zeroq} generate samples with the guidance of statistics hidden in BN layers, and subsequently use these generated images to fine-tune the quantized models to recover their accuracy. However, these methods are originally designed for CNN models, which cannot be directly applied to ViTs due to the absence of BN layers. 
To account for ViT models, Li \textit{et al.} \cite{li2022patch} propose PSAQ-ViT, which pioneers the study of data-free quantization on ViTs, introducing a metric called ``\textit{patch similarity}'' which is used for guiding the optimization of Gaussian noise images. PSAQ-ViT regularizes the image generation by adding an entropy maximization constraint to enrich the texture information of synthetic images. Later, Li \textit{et al.} \cite{li2023psaq} devise PSAQ-ViT v2, which improves the PSAQ-ViT by introducing a cyclic training framework that switches between generating training samples and training quantized models. Such a cyclic training framework largely improves the performance on both image classification and detection tasks. More recently, Ramachandran \textit{et al.} \cite{ramachandran2025CLAMP} propose CLAMP-ViT, which introduces a patch-level contrastive learning strategy to generate high-fidelity data, followed by a mixed-precision quantization schema. 

While PSAQ‑ViT v1~\cite{li2022patch} pioneered the use of patch‑similarity statistics for data‑free quantization of ViTs, our method differs from them in three respects. First, our RankQ-ViT differs from PSAQ-ViT in terms of the information metric. PSAQ-ViT maximizes entropy, a measure of distribution dispersion, which only captures the scatter of similarity values. As shown in Fig.~\ref{fig:psaq_vs_rankq}, RankQ-ViT instead directly promotes the rank of the similarity matrix, with a more fundamental indicator of the effective dimensionality of feature representations to encode richer structural information. 
Second, the optimization strategies are different. PSAQ optimizes synthetic samples by explicitly maximizing entropy, which uniformly applies a costly entropy estimation for all blocks during data generation. Our RankQ-ViT, in turn, uses a self-contrastive rank promotion mechanism to increase matrix rank by reducing patch correlations, supported by a theoretical lower-bound analysis, and a block-aware scheduling strategy that applies stronger promotion to shallow layers for better alignment with realistic images.
Third, our RankQ-ViT achieves superior performance to PSAQ‑ViT, which is detailed in Sec.~\ref{sec:exp}.



\section{Preliminaries}

\subsection{Architecture of Vision Transformers}
Let $X^{(l)} \in \mathbb{R}^{N \times D}$ denote the input tensor for the $l$-th transformer block, where $N$ represents the number of patches and $D$ represents the patch dimension, respectively. Then the $X^{(l)}$ is fed into the transformer block, which comprises two primary components: a multi-head self-attention (MHSA) mechanism, and a multilayer perceptron (MLP) module. This process can be formally defined:
\begin{equation}
    \hat{X}^{(l)} = {\rm{MHSA}}^{(l)}({\rm{LN}}(X^{(l)})) + X^{(l)},
\end{equation}
\begin{equation}
    X^{(l+1)} = {\rm MLP}({\rm LN}(\hat{X}^{(l)})) + \hat{X}^{(l)},
\end{equation}
where LN denotes the LayerNorm layer, and MHSA is:
\begin{equation}
    {\rm Attn}_i(Q_i, K_i, V_i) = {\rm Softmax}(Q_iK_i^T / \sqrt{D_h})V_i,
\end{equation}
\begin{equation}
    {\rm MHSA}(x) = {\rm concat}({\rm Attn}_1, {\rm Attn}_2, ..., {\rm Attn}_H)W^o,
\end{equation}
where $Q_i$, $K_i$, and $V_i$ denote the query, key, and value matrix transformed from input for the $i$-head. The $D_h$ denotes the embedding dimension for each head. The $H$ stands for the number of heads, and $W^o \in \mathbb{R}^{D_h \times D}$ is used for feature aggregation across heads. Subsequently, the MLP module encodes each patch feature independently, which is:
\begin{equation}
    {\rm MLP}(x) = {\rm FC}_2(\phi({\rm FC}_1(x))),
\end{equation}
where ${\rm FC}_1$ and ${\rm FC}_2$ are two consecutive linear layers, and $\phi(\cdot)$ represents the activation function.

\begin{table}[]
\centering
\caption{Key symbols in this work.}
\label{tab:symbols}
\small
\begin{tabular}{c|l}
\toprule
$X_G$ & generated image \\  
$Y$ & label for $X_G$ \\
$F_{X_G}^{(l)}$ & features in the $l$-th block \\
$S$ & patch similarity matrix \\
$\tau$ & temperature parameter \\
$\gamma$ & A small positive number \\
$D_p(\cdot)$ &dropout with probability $p$ \\
\bottomrule
\end{tabular}
\end{table}

\subsection{Data-free Quantization}
Mainstream data-free quantization methods~\cite{li2023psaq,choi2025mimiq,cai2020zeroq} mainly consist of two steps: data generation and model quantization, which are detailed below.
\subsubsection{Data Generation}
This step aims to generate synthetic samples for quantized model training. Specifically, the generated samples $X_G$ are treated as trainable parameters and optimized via the following objective:
\begin{equation}
    \min_{X_G}\mathcal{L}_{ target}(\mathcal{M}(X_G),Y) + \mathcal{L}_{prior}(X_G),
\end{equation}
where $\mathcal{M}$ and $Y$ denote the pre-trained model and the corresponding labels, respectively. 
The term $\mathcal{L}_{{target}}(\cdot, \cdot)$ is typically instantiated as the cross-entropy loss, while $\mathcal{L}_{{prior}}(\cdot)$ is often implemented using batch normalization (BN) statistics matching \cite{cai2020zeroq} or entropy regularization \cite{li2022patch}.

\subsubsection{Model Quantization}
In line with previous works \cite{li2022patch,ramachandran2025CLAMP}, we quantize all the weights and activations with the widely used uniform quantizer, which is defined as:
\begin{equation}
    {\rm Quant}(x_{\rm FP}, s, z, k) = {\rm clip}(\big \lfloor\frac{x_{\rm FP}}{s} - z\big \rceil, 0, 2^k - 1),
\end{equation}
where $x_{\rm FP}$, $s$, $z$, and $k$ denote the full-precision input, quantization scale, quantization zero-point, and bit-width, respectively. Consistent with common practice \cite{li2022patch, ramachandran2025CLAMP}, we employ channel-wise quantization for weights and layer-wise quantization for activations.

\begin{figure*}[!t] 
	\centering 
	\includegraphics[width=0.99\textwidth]{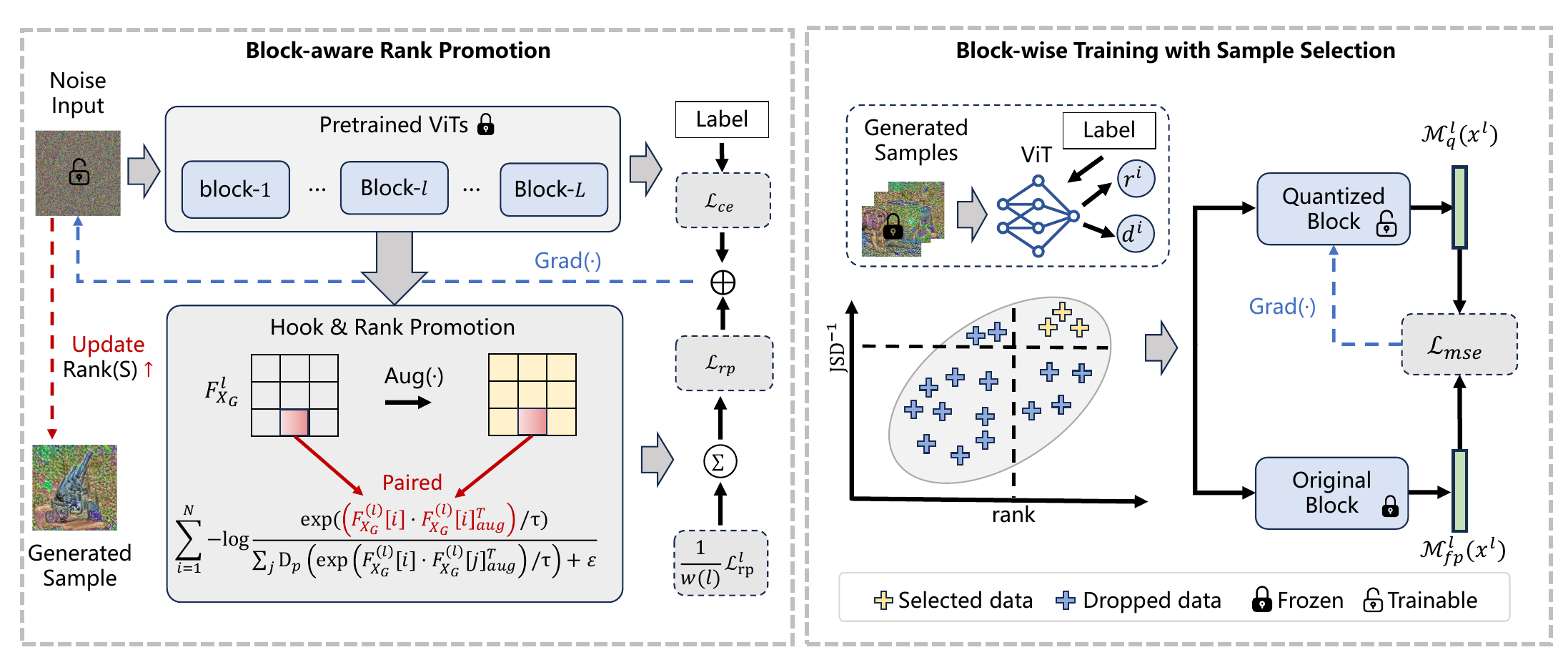} 
	\caption{Overview of RankQ-ViT. The workflow of RankQ-ViT involves two consecutive steps, namely block-aware rank promotion (the left panel) and block-wise training with sample selection (the right panel). First, the Gaussian noise images are encouraged to learn the high-rank similarity matrix by minimizing the $\mathcal{L}_{rp}$. Then, generated samples with higher ranks (see Eq.~\ref{eq:avg-rank}) and minimal Jensen-Shannon Divergence (JSD) (see Eq.~\ref{eq:jsd}) are selected to form the training set $\mathcal{D}$. Finally, the quantized model is forced to mimic the outputs of its full-precision version, recovering the accuracy degradation of quantization.}  \label{fig:overview} 
\end{figure*}

\section{Method}

\subsection{Block-aware Rank Promotion Mechanism}
\subsubsection{Motivation}

In this paper, we uncover an intriguing phenomenon related to the ranks of the patch similarity matrix. As shown in Fig.~\ref{fig:intro}, it is evident that realistic images tend to produce patch similarity matrices with higher ranks compared to Gaussian noise images. In matrix theory, a higher-rank matrix is typically associated with a greater amount of information, while a lower-ranked matrix contains less information.
In addition, we also observe that ranks significantly vary across network blocks, that is, ranks in the shallow blocks are generally higher than those of deeper ones. Therefore, according to the discussion above, the rank of the patch similarity matrix can be viewed as a metric to differentiate between the Gaussian noise images and realistic images, which can be further used to guide the sample generation, detailed in Fig.~\ref{fig:overview}. However, as indicated in Proposition~\ref{them:rank}, the training objective in \cite{li2022patch} cannot ensure high ranks during optimization, which motivates us to design new training paradigm to promote ranks effectively.

\subsubsection{Block-aware Rank Promotion Mechanism}
In this work, we propose a block-aware rank promotion mechanism to guide synthetic data generation. To be specific, let $X_G\in\mathbb{R}^{3 \times H \times W}$ denote the input image initialized by the Gaussian distribution, and $F_{X_G}^{(l)}\in\mathbb{R}^{N \times D}$ denote the features of $X_G$ after the $l$-th MHSA module. According to \cite{li2022patch}, the patch similarity matrix $S$ is defined as:

\begin{equation}
    S[i,j] = \frac{F_{X_G}^{(l)}[i,:]\cdot F_{X_G}^{(l)}[j,:]^{\rm T}}{||F_{X_G}^{(l)}[i,:]|| \cdot ||F_{X_G}^{(l)}[j,:]|| },
\end{equation}
where $F_{X_G}^{(l)}[i,:]$ and $F_{X_G}^{(l)}[j,:]$ denote the $i$-th and $j$-th patch feature of $F_{X_G}^{(l)}$, respectively. 

As shown in Fig.~\ref{fig:intro}, since the Gaussian noise contains less information, the rank of $S$ (denoted as Rank($S$) for short) is lower than the realistic images. Therefore, the key question lies in \textit{how to increase the Rank($S$) efficiently}. A straightforward approach is to leverage advanced deep learning frameworks (e.g., PyTorch) or Singular Value Decomposition (SVD) to compute the ranks, followed by optimization techniques to maximize Rank($S$). Unfortunately, these methods are often computationally expensive, as they typically involve extensive matrix computations (e.g., matrix inversion) at every iteration and block, leading to significant time overhead.


To remedy this, we introduce a rank promotion method with self-contrastive learning \cite{oord2018representation} on a patch level, which is:
\begin{equation} \label{eq:rpl}
    \mathcal{L}_{rp}^l = \sum_{i=1}^{N} {\rm -log} \frac{\exp((F_{X_G}^{(l)}[i,:]\cdot F_{X_G}^{(l)}[i,:]^{\rm T}_{aug})/\tau)}{\sum_{j}{\rm D}_p(\exp((F_{X_G}^{(l)}[i,:]\cdot F_{X_G}^{(l)}[j,:]^{\rm T}_{aug})/ \tau)) + \gamma},
\end{equation}
where ${\rm D}_p(\cdot)$ denotes the dropout operation with probability $p$, which is set to $0.5$ in our experiments to achieve diversity. $\tau$ is the temperature parameter to control the smoothness of similarity, which is set to 0.1. The constant $\gamma$ is set to 0.1 to avoid division by zero. $F_{X_G}^{(l)}[i,:]^{\rm T}_{aug}$ denotes the features extracted from the augmented images (e.g., those processed by image rolling). 


Unlike the optimization strategy use in PSAQ-ViT, which captures only the dispersion of similarity values and does not ensure coverage of diverse feature directions, potentially limiting the representational diversity of synthetic samples and their effectiveness for quantization calibration. Our method, in turn, by minimizing Eq.~\ref{eq:rpl}, each patch is forced to learn distinctive features, achieving low cosine similarity between different patches, which effectively increases the rank of the patch similarity matrix (see Proposition~\ref{them:1}).


\begin{prop} \label{them:rank}
A high entropy in PSAQ~\cite{li2022patch} does not necessarily imply a high-rank similarity matrix. 
Specifically, for a small positive integer $r$, there exists a similarity matrix $S$ such that $\operatorname{Rank}(S)=r$, while their entropy remains non-trivial.
\end{prop}
\begin{proof}
 Please refer to the Appendix.
\end{proof}

\begin{prop} \label{them:1}
Let $x_i$ is normalized patch feature, and $X=[x_1,\dots,x_N]^\top \in \mathbb{R}^{N\times D}$ be a matrix whose rows are normalized such that $\|x_i\|_2=1$. 
Assume the pairwise cosine similarities satisfy
\begin{equation}
|x_i^\top x_j| \le \epsilon, \quad \forall i\neq j.
\end{equation}
Then the rank of $X$ is lower bounded by
\begin{equation}
\mathrm{rank}(X) \ge \frac{N}{1+(N-1)\epsilon^2}.
\end{equation}
\end{prop}

\begin{proof}
 Please refer to the Appendix.
\end{proof}

Moreover, as shown in Fig.~\ref{fig:intro}, we observe that the ranks decrease when the model goes deeper for realistic images. Therefore, we force the generated images to mimic this behavior by imposing a smaller weight for rank promotion loss on deeper blocks. Finally, we sum up the rank promotion loss across all blocks to formulate the overall loss:
\begin{equation}
    \mathcal{L}_{rp} = \sum_{l=1}^L \frac{1}{w(l)}\mathcal{L}_{rp}^l,
\end{equation}
where $L$ is the total number of blocks, $w(l) = \frac{l}{L}$ is a weight factor for the $l$-th ViT block, which prioritizes the shallow layers to learn more distinct feature representations. To stabilize the training process, in the early optimization iterations, we only use the standard cross-entropy loss $\mathcal{L}_{ce}$ to initialize the semantic information. After that, the rank promotion loss is introduced, and the total loss becomes:
\begin{equation} \label{eq:total}
    \mathcal{L}_{all} = \mathcal{L}_{rp} + \alpha \mathcal{L}_{ce},
\end{equation}
where $\alpha$ is set to 1. In this stage, all the model parameters are frozen, only the generated images are updated by back-propagation.


\subsection{Block-wise Training with Sample Selection}
Till now, we have obtained a handful of generated samples by optimizing random noise images according to Eq.~\ref{eq:total}. However, as highlighted in previous studies \cite{fan2024data}, the reliability of these generated samples can be compromised due to the inherent instability of the generation process. As a consequence, using all generated samples indiscriminately may adversely affect the learning of quantized models. To address this issue, we propose a sample selection strategy aimed to identify reliable samples, facilitating subsequent reconstruction procedures.
Specifically, we first compute the averaged rank of the patch similarity matrix across blocks for each sample, defined as:
\begin{equation} \label{eq:avg-rank}
    r_i = \frac{1}{L}\sum_{l=1}^{L}{\rm Rank}(S_l),
\end{equation}
where $S_l$ is the patch similarity matrix of the $l$-th block. As mentioned previously, the rank serves as a measure of the fidelity of a generated sample, and the higher ranks generally imply better image fidelity. Furthermore, to incorporate semantic information, we propose to use the Jensen-Shannon divergence (JSD) \cite{englesson2021generalized} to evaluate the semantic consistency between the one-hot label $\textbf{y}_i$ and the predicted probability distribution $\textbf{p}_i$. This is formulated as:
\begin{equation}  \label{eq:jsd}
\begin{aligned}
     d_i  = {\rm JSD}_\pi(\textbf{y}_i, \textbf{p}_i) &= \pi \cdot {\rm KL}(\textbf{y}_i || m) + (1 - \pi) \cdot {\rm KL} (\textbf{p}_i || m), \\
    m & = \pi\cdot\textbf{y}_i + (1-\pi)\cdot\textbf{p}_i,
\end{aligned}
\end{equation}
where ${\rm KL} (\cdot)$ denotes the KL-Divergence, and $\pi$ is set to 0.5. 

Finally, we pick samples with smaller JSD and higher ranks for the following adopted fine-tuning process. In this stage, a block-wise reconstruction paradigm \cite{librecq,jiang2025aiqvit} is adopted:
\begin{equation} \label{eq:ft}
\begin{aligned}
    \mathcal{L}_{ft} & = \frac{1}{|\mathcal{D}|}\sum_{X\in\mathcal{D}}||\mathcal{M}^{l}_q(X^{l}) - \mathcal{M}^{l}_{\rm FP}(X^{l})||_{\rm F}, \\
    \mathcal{D} & = {\rm arg max}_{|\hat{\mathcal{D}}| \leq N_{sel}}\sum_{X_i \in \hat{\mathcal{D}}} r_{X_i} -d_{X_i},
    \end{aligned}
\end{equation}
where $|\mathcal{D}|$ denotes the size of $\mathcal{D}$, $\mathcal{M}^{l}_q(\cdot)$ and $\mathcal{M}^{l}_{\rm FP}(\cdot)$ represent the outputs of $l$-th quantized block and full-precision block, respectively. $N_{sel}$ is the number of selected samples. Here, we omit the normalization procedure of $d_{X_i}$ and $r_{X_i}$ for simplicity.  $X^l$ denotes the input for the $l$-th block, and the $||\cdot||_{\rm F}$ represents the Frobenius norm. 

\begin{figure}[!t] 
	\centering 
	\includegraphics[width=0.49\textwidth]{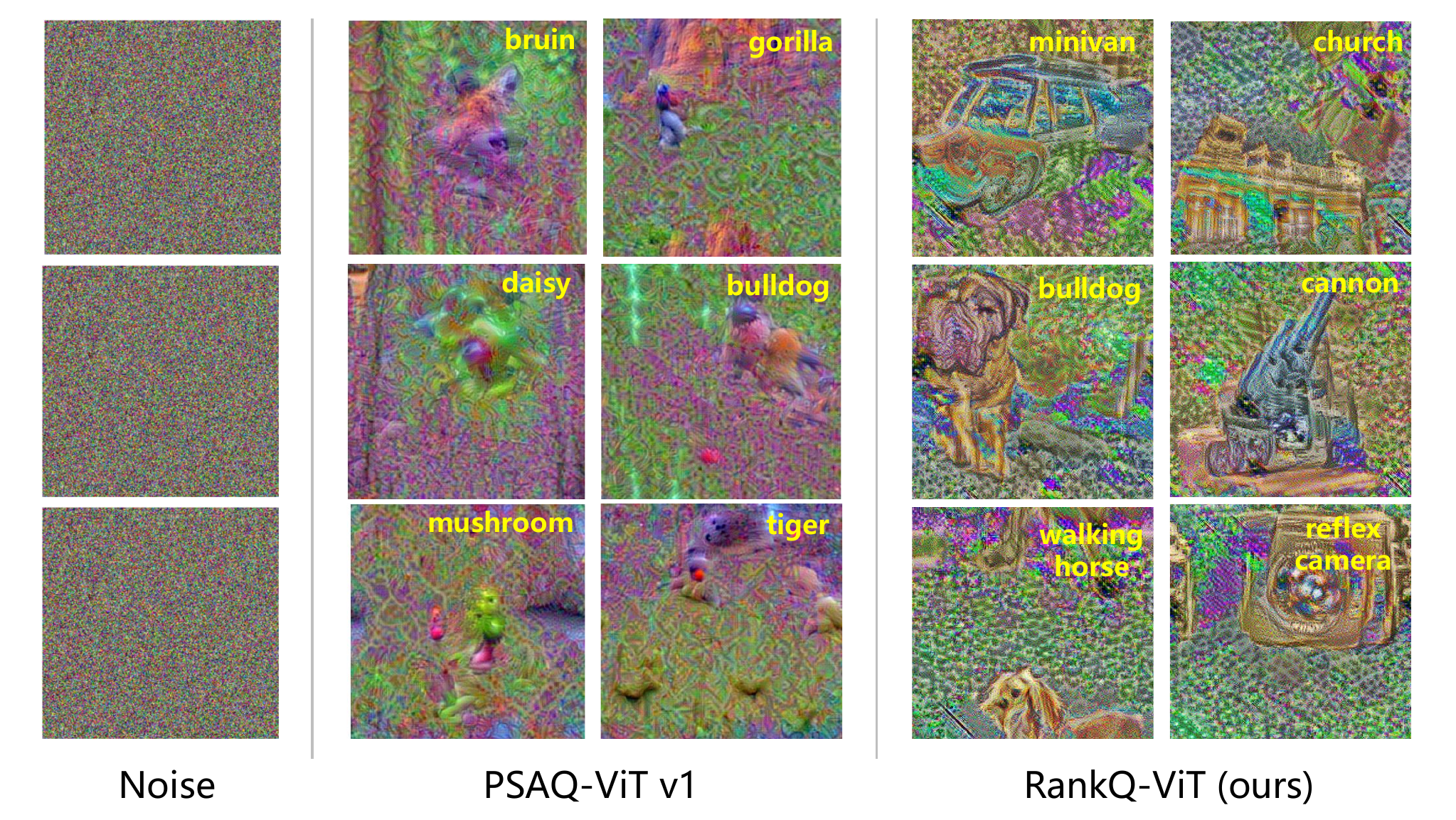} 
 	\caption{The synthetic samples (224×224 pixels) shown in the figure are generated by PSAQ-ViT v1 and our proposed RankQ-ViT. The label associated with each generated image is indicated in the upper-right corner. The results demonstrate that images produced by PSAQ-ViT contain obscure semantic information, whereas those generated by our RankQ-ViT exhibit significantly stronger semantic content. Furthermore, the images from RankQ-ViT show richer texture details compared to those from PSAQ-ViT.} \label{fig:syn-image}
\end{figure}

\section{Experiments} \label{sec:exp}
\subsection{Experimental Setups}
\begin{table*}[!t]
\centering
\caption{Quantization results on the ImageNet dataset with W4/A8 and W8/A8 quantization. ``No Data'' means that no realistic images are used. ``Data Num.'' means the number of images used for calibration. ``Precision'' denotes the bit-width of weights and activations. The top-1 accuracy (\%) is reported as the evaluation metric.} \label{tab:imgnet}
\renewcommand{\arraystretch}{1.0}
\setlength{\tabcolsep}{11pt}
\begin{tabular}{cccccccc} 
\toprule
Model                   & Method         & No Data & Data Num. & Precision   & Top-1 Acc. (\%) & Precision   & Top-1 Acc. (\%) \\ \midrule
\multirow{5}{*}{ViT-S}  & Full-Precision & -        & -         & W32/A32   & 81.39     & W32/A32   & 81.39     \\
& Standard \cite{li2022patch} &$\times$ &32 &W4/A8 &19.91 &W8/A8 &30.28\\
                        & PSAQ-ViT v1 \cite{li2022patch}    &$\checkmark$         & 32        & W4/A8   & 20.84     & W8/A8   & 31.45     \\
                        & PSAQ-ViT v2 \cite{li2023psaq} &$\checkmark$ &- &W4/A8 &77.84 &W8/A8 &81.05\\ 
                        & RepQ-ViT \cite{li2023repq}      &$\times$         & 32        & W4/A8   &75.60           & W8/A8   &81.17           \\
                        &  RankQ-ViT (ours)          &$\checkmark$         & 32        & W4/A8   & \textbf{78.76}$_{\pm0.12}$     & W8/A8   & \textbf{81.19}$_{\pm0.08}$     \\
                        \midrule
\multirow{6}{*}{ViT-B}  & Full-Precision &-         & -         & W32/A32   & 84.53     & W32/A32   & 84.53     \\
& Standard\cite{li2022patch} &$\times$ &32 &W4/A8 &24.76 &W8/A8 &36.65 \\
                        & PSAQ-ViT v1 \cite{li2022patch}   &$\checkmark$         & 32        & W4/A8   & 25.31     & W8/A8   & 37.36     \\
                        & PSAQ-ViT v2 \cite{li2023psaq} &$\checkmark$ &- & W4/A8 &81.64 &W8/A8 &84.01\\
                        & RepQ-ViT \cite{li2023repq}      &$\times$         & 32        & W4/A8   & 76.29     & W8/A8   & 81.45     \\
                        & CLAMP-ViT \cite{ramachandran2025CLAMP}      &$\checkmark$         & 32        & W4/A8   &78.73           & W8/A8   &84.19           \\
                        &  RankQ-ViT (ours)           &$\checkmark$         & 32        & W4/A8   &\textbf{82.44}$_{\pm0.14}$           & W8/A8   &\textbf{84.25}$_{\pm0.07}$           \\
                        \midrule 
                        \multirow{7}{*}{DeiT-T}  & Full-Precision & -        & -         & W32/A32   & 72.21     & W32/A32   & 72.21     \\
                        & Standard \cite{li2022patch} &$\times$ &32 &W4/A8 &65.20 &W8/A8 &71.27 \\
                        & PSAQ-ViT v1 \cite{li2022patch}    &$\checkmark$         & 32        & W4/A8   & 65.57     & W8/A8   & 71.56     \\
                        & PSAQ-ViT v2 \cite{li2023psaq}    &$\checkmark$         & 32        & W4/A8   & 68.61     & W8/A8   & {72.17}     \\
                        
                        & RepQ-ViT \cite{li2023repq}       &$\times$         & 32        & W4/A8   &67.40
           & W8/A8   &71.94           \\
           & DFQ-ViT \cite{tong2025dfq} &$\checkmark$ &16 & W4/A8 &65.84 &W8/A8 &71.64\\
           & CLAMP-ViT \cite{ramachandran2025CLAMP}       &$\checkmark$         & 32        & W4/A8   &\textbf{69.93}
           & W8/A8   &72.17           \\
                        & RankQ-ViT (ours)           &$\checkmark$         & 32        & W4/A8   & {68.55}$_{\pm0.12}$     & W8/A8   & \textbf{72.22}$_{\pm0.06}$     \\
                        \midrule
\multirow{8}{*}{DeiT-S} & Full-Precision &-         & -         & W32/A32   & 79.85     & W32/A32   & 79.85     \\
& Standard \cite{li2022patch} &$\times$ &32 &W4/A8 &72.10 &W8/A8 &76.00 \\
                        & PSAQ-ViT v1 \cite{li2022patch}    & $\checkmark$        & 32        & W4/A8   &73.23           & W8/A8   &76.92           \\
                        & PSAQ-ViT v2 \cite{li2023psaq}   &$\checkmark
                        $         & -         & W4/A8   &76.36           & W8/A8   &79.56           \\
                        & RepQ-ViT \cite{li2023repq}      &$\times$         & 32        & W4/A8   & 75.19          & W8/A8   & 79.72          \\
                        & DFQ-ViT \cite{tong2025dfq} &$\checkmark$ &16 & W4/A8 &73.69 &W8/A8 &77.04\\
                        & CLAMP-ViT \cite{ramachandran2025CLAMP}     &$\checkmark$         & 32        & W4/A8   &77.03           & W8/A8   &{79.55}           \\
                        &  RankQ-ViT (ours)           &$\checkmark$         & 32        & W4/A8   &\textbf{77.86}$_{\pm0.15}$           & W8/A8   &\textbf{79.59}$_{\pm0.05}$           \\
                        \midrule
                        \multirow{7}{*}{DeiT-B} & Full-Precision &-         & -         & W32/A32   & 81.80     & W32/A32   & 81.80     \\
                        & Standard \cite{li2022patch} &$\times$ &32 &W4/A8 &76.25 &W8/A8 &78.61 \\
                        & PSAQ-ViT v1 \cite{li2022patch} & $\checkmark$        & 32        & W4/A8   &77.05           & W8/A8   &79.10           \\
                        & PSAQ-ViT v2 \cite{li2023psaq}   &$\checkmark
                        $         & -         & W4/A8   &79.49           & W8/A8   &81.52           \\
                        & RepQ-ViT \cite{li2023repq}      &$\times$         & 32        & W4/A8   &77.16           & W8/A8   &80.50           \\
                        & DFQ-ViT \cite{tong2025dfq} &$\checkmark$ &16 & W4/A8 &77.23 &W8/A8 &79.51\\
                        &  RankQ-ViT (ours)           &$\checkmark$         & 32        & W4/A8   &\textbf{80.63}$_{\pm0.10}$           & W8/A8   &\textbf{81.63}$_{\pm0.04}$           \\
                        \midrule
                        
\multirow{9}{*}{Swin-T} & Full-Precision &-         & -         & W32/A32 & 81.35     & W32/A32   & 81.35     \\
& Standard \cite{li2022patch}&$\times$ &32 &W4/A8 &70.16 &W8/A8 &74.22 \\
                        & PSAQ-ViT v1 \cite{li2022patch}    &$\checkmark$         & 32        & W4/A8   &71.79           & W8/A8   &75.35           \\
                        & PSAQ-ViT v2 \cite{li2023psaq}   &$\checkmark$         & -         & W4/A8   & 76.28          & W8/A8   &80.21           \\
                        & RepQ-ViT \cite{li2023repq}      &$\times$         & 32        & W4/A8   & 72.92          & W8/A8   &\textbf{81.20}           \\
                        & DFQ-ViT \cite{tong2025dfq} &$\checkmark$ &16 & W4/A8 &74.01 &W8/A8 &76.85\\
                        & RIEM \cite{pei2024data} &$\checkmark$ &- &W4/A8 &76.28 &W8/A8 &80.85\\ 
                        & CLAMP-ViT \cite{ramachandran2025CLAMP}     &$\checkmark$         & 32        & W4/A8   & \textbf{80.28}          & W8/A8   &  81.17         \\
                        & RankQ-ViT (ours)           & $\checkmark$        & 32        & W4/A8   &79.24$_{\pm0.10}$           & W8/A8   & 81.14$_{\pm0.05}$          \\
                        \midrule 
                    \multirow{9}{*}{Swin-S}    & Full-Precision & -        & -         & W32/A32 & 83.20     & W32/A32 & 83.20     \\
                    & Standard \cite{li2022patch} &$\times$ &32 &W4/A8 &73.33 &W8/A8 &75.19 \\
                  & PSAQ-ViT v1 \cite{li2022patch}  &$\checkmark$         & 32        & W4/A8   &75.14           & W8/A8   &76.64           \\
                        & PSAQ-ViT v2 \cite{li2023psaq}   & $\checkmark$        & -         & W4/A8   &78.86           & W8/A8   &82.13           \\
                        & RepQ-ViT \cite{li2023repq}      &$\times$         & 32        & W4/A8   &81.31           & W8/A8   &83.02           \\
                        & DFQ-ViT \cite{tong2025dfq} &$\checkmark$ &16 & W4/A8 &76.24 &W8/A8 &77.21\\
                         & RIEM \cite{pei2024data}&$\checkmark$ &- &W4/A8 &79.84 &W8/A8 &82.96\\ 
                        & CLAMP-ViT \cite{ramachandran2025CLAMP}      & $\checkmark$        & 32        & W4/A8   & \textbf{82.51}          & W8/A8   &82.57           \\
                        &  RankQ-ViT (ours)           &$\checkmark$         & 32        & W4/A8   &82.36$_{\pm0.15}$           & W8/A8   &  \textbf{83.19}$_{\pm0.09}$         \\ \bottomrule
\end{tabular}
\end{table*}

\begin{table*}[!t]
\centering
 \caption{Quantization results on the ImageNet dataset with W4/A4 and W5/A5 quantization.} \label{tab:imgnet-2}
\renewcommand{\arraystretch}{1.0}
\setlength{\tabcolsep}{11pt}
\begin{tabular}{cccccccc} 
\toprule
Model                   & Method         & No Data & Data Num. & Precision   & Top-1 Acc. (\%) & Precision   & Top-1 Acc. (\%) \\ \midrule
\multirow{5}{*}{ViT-S}  & Full-Precision & -        & -         & W32/A32   & 81.39     & W32/A32   & 81.39     \\
& Standard &$\times$ &32 &W4/A4 &39.42 &W5/A5 &71.78\\
                        & PSAQ-ViT v1 \cite{li2022patch}    &$\checkmark$         & 32        & W4/A4   &0.12     & W5/A5   &24.78     \\
                        & PSAQ-ViT v2 \cite{li2023psaq} &$\checkmark$ &- & W4/A4   & 9.78    & W5/A5   &60.12\\ 
                       
                        &  RankQ-ViT (ours)          &$\checkmark$         & 32        & W4/A4   &\textbf{44.06}$_{\pm0.18}$     & W5/A5   &\textbf{68.07}$_{\pm0.17}$    \\
                        \midrule
\multirow{5}{*}{ViT-B}  & Full-Precision &-         & -         & W32/A32   & 84.53     & W32/A32   & 84.53     \\
& Standard &$\times$ &32 &W4/A4 &14.80 &W5/A5 &70.92 \\
                        & PSAQ-ViT v1 \cite{li2022patch}   &$\checkmark$         & 32        & W4/A4   & 0.61     & W5/A5   & 14.78     \\
                        & PSAQ-ViT v2 \cite{li2023psaq} &$\checkmark$ &-& W4/A4   & 6.30     & W5/A5   &48.78 \\
                        &  RankQ-ViT (ours)           &$\checkmark$         & 32        & W4/A4   &\textbf{17.77}$_{\pm0.22}$           & W5/A5   &{\textbf{50.19}}$_{\pm0.24}$           \\
                        \midrule 
                        \multirow{5}{*}{DeiT-T}  & Full-Precision & -        & -         & W32/A32   & 72.21     & W32/A32   & 72.21     \\
                        & Standard &$\times$ &32 &W4/A4 &31.79 &W5/A5 &60.42 \\
                        & PSAQ-ViT v1 \cite{li2022patch}    &$\checkmark$         & 32        &W4/A4 &20.61 &W5/A5 &52.19    \\
                        & PSAQ-ViT v2 \cite{li2023psaq}    &$\checkmark$         & 32        &W4/A4 &20.20 &W5/A5 &52.18     \\

                        & RankQ-ViT (ours)           &$\checkmark$         & 32        & W4/A4   & \textbf{{44.35}}$_{\pm0.20}$     & W5/A5   & \textbf{62.38}$_{\pm0.14}$     \\
                        \midrule
\multirow{5}{*}{DeiT-S} & Full-Precision &-         & -         & W32/A32   & 79.85     & W32/A32   & 79.85     \\
&Standard$^\dagger$ &$\times$ &32 &W4/A4 &29.23 &W5/A5 &61.54 \\
                        & PSAQ-ViT v1 \cite{li2022patch}    & $\checkmark$        & 32        & W4/A4   &6.23           & W5/A5   &48.26           \\
                        & PSAQ-ViT v2 \cite{li2023psaq}   &$\checkmark
                        $         & -         & W4/A4   &34.98           & W5/A5   &63.21           \\
                       
                        &  RankQ-ViT (ours)           &$\checkmark$         & 32        & W4/A4   &\textbf{50.40}$_{\pm0.27}$           & W5/A5   &\textbf{64.13}$_{\pm0.19}$           \\
                        \midrule
                        \multirow{5}{*}{DeiT-B} & Full-Precision &-         & -         & W32/A32   & 81.80     & W32/A32   & 81.80     \\
                        & Standard &$\times$ &32 &W4/A4 &64.06 &W5/A5 &77.29 \\
                        & PSAQ-ViT v1 \cite{li2022patch} & $\checkmark$        & 32        & W4/A4   &10.26           & W5/A5   &55.45           \\
                        & PSAQ-ViT v2 \cite{li2023psaq}   &$\checkmark
                        $         & -         & W4/A4   &45.01           & W5/A5   &70.23           \\

                        &  RankQ-ViT (ours)           &$\checkmark$         & 32        & W4/A4   &\textbf{64.34}$_{\pm0.17}$           & W5/A5   &\textbf{75.16}$_{\pm0.21}$           \\
                        \midrule
                        
\multirow{5}{*}{Swin-T} & Full-Precision &-         & -         & W32/A32 & 81.35     & W32/A32   & 81.35     \\
& Standard &$\times$ &32 &W4/A4 &33.59 &W5/A5 &74.11 \\
                        & PSAQ-ViT v1 \cite{li2022patch}    &$\checkmark$         & 32        &W4/A4 &22.51 &W5/A5 &57.78          \\
                        & PSAQ-ViT v2 \cite{li2023psaq}   &$\checkmark$         & -        &W4/A4 &48.67 &W5/A5 &67.67          \\
                        & RankQ-ViT (ours)           & $\checkmark$        & 32        &W4/A4 &\textbf{71.57}$_{\pm0.22}$ &W5/A5 &\textbf{79.05}$_{\pm0.20}$        \\
                        \midrule 
                    \multirow{5}{*}{Swin-S}    & Full-Precision & -        & -         & W32/A32 & 83.20     & W32/A32 & 83.20     \\
                    & Standard&$\times$ &32 &W4/A4 &78.71 &W5/A5 &81.64 \\
                  & PSAQ-ViT v1 \cite{li2022patch}  &$\checkmark$         & 32        & W4/A4   &12.09           & W5/A5   &77.01           \\
                        & PSAQ-ViT v2 \cite{li2023psaq}   & $\checkmark$        & -         & W4/A4   &37.89           & W5/A5   &78.37           \\
    
                        &  RankQ-ViT (ours)           &$\checkmark$         & 32        & W4/A4   &\textbf{77.78}$_{\pm0.23}$           & W5/A5   &  \textbf{81.25}$_{\pm0.26}$         \\ \bottomrule
\end{tabular}
\end{table*}

\subsubsection{Datasets and Models}
We utilize the ImageNet \cite{krizhevsky2012imagenet}, COCO~\cite{lin2014microsoft}, ModelNet40 \cite{wu20153d}, and ShapeNet \cite{yi2016scalable} benchmarks to evaluate the performance of our method on four computer vision tasks (including image classification, object detection, instance segmentation, and point cloud classification).
Besides, to assess the efficacy of our RankQ-ViT, we employ a diverse range of Transformer architectures, including ViT \cite{dosovitskiy2020image}, DeiT \cite{touvron2021training}, Swin Transformer \cite{liu2021swin}, and Point Transformer \cite{zhao2021point}.

\subsubsection{Experimental Details}
For the image classification task, we generate 128 synthetic samples with resolutions of $224 \times 224$, and then select 32 samples from these with the sample selection strategy defined in Eq.~\ref{eq:ft} for the following optimization. We then perform a block-wise fine-tuning paradigm with a batch size of 32 and train for 2,000 iterations, employing the Adam optimizer with an initial learning rate of $4\times10^{-4}$ for all the network variants. Specifically, we select Standard \cite{li2022patch}, PSAQ-ViT v1 \cite{li2022patch}, PSAQ-ViT v2 \cite{li2023psaq}, CLAMP-ViT \cite{ramachandran2025CLAMP}, BRECQ \cite{librecq}, RepQ-ViT \cite{li2023repq}, DFQ-ViT \cite{tong2025dfq} and RIEM \cite{pei2024data} as competing methods. For the object detection and instance segmentation task, we generate 128 synthetic samples and then select 32 samples for calibration. We then perform a block-wise fine-tuning paradigm with a batch size of 1 and train for 2,000 iterations, employing the Adam optimizer with an initial learning rate of $2\times10^{-4}$ for all the network variants. All experiments are implemented in PyTorch and evaluated on a single NVIDIA RTX-4090 GPU with 24GB memory. The system runs on Ubuntu 22.04, with a total memory of 64GB.

As for the point cloud classification task, we also generate 128 point clouds and then select 32 point clouds for block-wise optimization, each comprising 1,024 points. During block-wise fine-tuning, each block is trained for 2,000 iterations using the Adam optimizer with an initial learning rate of $1\times10^{-4}$. The batch size is set to 32 for both ModelNet40 and ShapeNet in our experiments. Note that, since there is no published work on data-free quantization for 3D point cloud understanding, we compare our method with quantized models calibrated using realistic point clouds and Gaussian point clouds. Additionally, we also reproduce the RepQ-ViT \cite{li2023repq}, BRECQ \cite{librecq}, PSAQ-ViT v1 \cite{li2022patch}, PSAQ-ViT v2 \cite{li2023psaq}, and AdaRound \cite{nagel2020up} for the point cloud classification task using their official code for comparison.

\begin{table*}[t]
\centering
\caption{Quantization results of point cloud classification.  ``mACC'' and ``OA'' denote the mean of class-wise accuracy (\%) and the
overall accuracy (\%), respectively.} \label{tab:pc}
\begin{tabular}{c|ccccccccc}
\toprule
Datasets                    & Method     &No Data    & Data Num. & Precision   & mAcc (\%) & OA (\%)  & Precision   & mAcc (\%) & OA (\%)              \\
\midrule
\multirow{10}{*}{ModelNet40} & Full-Precision  & - &-         & W32/A32 &83.99 &88.00 &W32/A32 &83.99 &88.00                      \\
                            & Gaussian    &$\checkmark$   & 32         & W4/A8   &77.36 &79.16 &W8/A8 &82.01 &82.50                      \\
                            & Real      &$\times$     & 32   &W4/A8      &80.00  &84.51    &W8/A8  &\textbf{83.67} &87.18                     \\
                            & Baseline &$\checkmark$ &32 &W4/A8 &79.85 &83.98 &W8/A8 &83.26 &87.00\\
                            & PSAQ-ViT v1 \cite{li2022patch} &$\checkmark$ &32 &W4/A8 &79.60 &83.65 &W8/A8 &83.21 &87.16\\
                            & PSAQ-ViT v2 \cite{li2023psaq} &$\checkmark$ &- &W4/A8 &80.00 &84.65 &W8/A8 &83.28 &87.01\\
                            & RepQ-ViT \cite{li2023repq}    &$\times$      & 32         & W4/A8   &79.98 &84.78 &W8/A8 &83.47 &86.95                     \\
                            &BRECQ \cite{librecq} &$\times$      & 32 &W4/A8 &79.78 &84.65 &W8/A8 &83.43 &87.09\\ 
                            & AdaRound \cite{nagel2020up} &$\times$      & 32 &W4/A8 &77.85 &83.26 &W8/A8 &82.64 &86.00\\
                            & RankQ-ViT (ours)   &$\checkmark$        & 32         & W4/A8   &\textbf{80.24}$_{\pm0.24}$ & \textbf{85.27}$_{\pm0.15}$ & W8/A8 &83.55$_{\pm0.07}$ &\textbf{87.40}$_{\pm0.08}$                   \\
                            \midrule
\multirow{10}{*}{ShapeNet}   & Full-Precision & -  &-        & W32/A32 &93.97 &98.16 & W32/A32 &93.97 &98.16                      \\
                            & Gaussian   &$\checkmark$    & 32         & W4/A8   &91.26 &95.78  &W8/A8 &92.71 &96.08 \\
                            & Real  & $\times$        & 32         & W4/A8   &93.11 &97.53 &W8/A8 &\textbf{95.51} &97.87 \\
                            & Baseline & $\checkmark$        & 32         & W4/A8 &92.51 &96.02 &W8/A8 &95.00 &97.21\\
                            & PSAQ-ViT v1 \cite{li2022patch} & $\checkmark$        & 32 &W4/A8 &92.00 &96.15 &W8/A8 &95.11 &97.18  \\
                            & PSAQ-ViT v2 \cite{li2023psaq} &$\checkmark$        & 32 &W4/A8 &92.37 &96.19 &W8/A8 &94.87 &97.03\\
                            & BRECQ \cite{librecq} &$\times$ &32 &W4/A8 &92.67 &97.78 &W8/A8 &95.03 &97.18\\
                            & AdaRound \cite{nagel2020up} &$\times$ &32 & W4/A8 &91.88 &96.87 &W8/A8 &94.48 &96.89\\
                            & RepQ-ViT \cite{li2023repq}  & $\times$         & 32         & W4/A8   & 92.88 &97.03  & W8/A8 &95.01 &97.22                    \\
                            &  RankQ-ViT (ours)      &$\checkmark$     & 32         & W4/A8   &\textbf{93.16}$_{\pm0.16}$ &\textbf{97.81}$_{\pm0.10}$ &W8/A8 &{95.43}$_{\pm0.08}$ &\textbf{97.95}$_{\pm0.10}$ \\
                            \bottomrule
\end{tabular}
\end{table*}

\subsection{Competing with the State-of-the-art Methods}
\subsubsection{Results on Image Classification}
As shown in Tab.~\ref{tab:imgnet}, RankQ-ViT consistently outperforms other competing methods in most quantization settings, demonstrating its effectiveness in preserving the accuracy of ViT models under various quantization scenarios. Notably, our method surpasses CLAMP-ViT by a significant margin of 3.71\% on ViT-B with W4/A8 quantization, primarily attributed to the effectiveness of our rank promotion mechanism and sample selection strategy. These results demonstrate the effectiveness of carefully designed quantization techniques in preserving the accuracy of ViT models. Furthermore, for most quantization cases, our RankQ-ViT closely approximates the results of full-precision models, indicating its potential for real-world applications where model accuracy and efficiency are crucial.
Notably, our RankQ-ViT outperforms the Full-Precision model on DeiT-T with W8/A8 quantization. Meanwhile, PSAQ-ViT v1 consistently exhibits the poorest performance in most cases, likely caused by the absence of fine-tuning, which results in inaccurate quantization parameters. 
In addition, CLAMP-ViT performs better than RankQ-ViT on Swin-T, since it uses the cyclically adaptive data generation process, which requires a significantly longer runtime, as suggested by \cite{ramachandran2025CLAMP}. 


We also evaluate our methods on more challenging quantization regimes (i.e., W4/A4 and W5/A5), which are shown in Tab.\ref{tab:imgnet-2}. Specifically, as for W5/A5 quantization, the PSAQ-ViT v1 method struggles due to the lack of training. In addition, the results suggest that most competing methods lose their ability in W4/A4 quantization. For instance, the PSAQ-ViT v1 and PSAQ-ViT v2 only achieve 0.12\% and 9.78\% accuracy on ViT-S quantization, which is far from the practical applications. Note that Standard obtains the best performance in some cases (i.e., ViT-B with W5/A5 quantization), but it still requires additional realistic images for calibration. Instead, our method generally outperforms the competing methods by a large margin in a data-free fashion.

\begin{table}[!bhpt]
\centering
\caption{Quantization results on COCO dataset. ``$\text{AP}^{\text{box}}$'' means the box average precision for object detection, and ``$\text{AP}^{\text{mask}}$'' means the mask average precision for instance segmentation.} \label{tab:coco}
\begin{tabular}{cccccc}
\toprule
Model                                                                                  & Method         & Prec. & AP$_{\rm box}$   & AP$_{\rm mask}$   \\
\midrule
\multirow{5}{*}{\begin{tabular}[c]{@{}l@{}}Swin-S\\ Cascade\\ Mask R-CNN\end{tabular}} & FP                 & W32/A32 & 51.9 & 48.0 \\
                & Standard       & W4/A8   & 47.3 & 40.9 \\
                & Gaussian  & W4/A8   & 38.1 & 35.2 \\
                & PSAQ v1 \cite{li2022patch} &W4/A8 &42.6 &36.9\\
                & PSAQ v2 \cite{li2023psaq}  & W4/A8   & 47.9 & 41.4 \\
                & RankQ-ViT        & W4/A8   & \textbf{48.3}$_{\pm0.02}$ & \textbf{41.7}$_{\pm0.01}$ \\
            \midrule
\multirow{5}{*}{\begin{tabular}[c]{@{}l@{}}Swin-S\\ Mask R-CNN\end{tabular}}           &  FP                  & W32/A32 & 48.5 & 43.3 \\
              & Standard        & W4/A8   & 44.3 & 37.1 \\
              & Gaussian   & W4/A8   & 37.3 & 30.2 \\
              & PSAQ v1 \cite{li2022patch} &W4/A8 &40.9 &32.9\\
            & PSAQ v2 \cite{li2023psaq} & W4/A8   & 44.9 & 37.8 \\
            & RankQ-ViT       & W4/A8   & \textbf{45.1}$_{\pm0.02}$ & \textbf{37.8}$_{\pm0.01}$ \\
            \bottomrule
\end{tabular}
\end{table}

\subsubsection{Results on Point Cloud Classification}
To further demonstrate the effectiveness of RankQ-ViT, we select Point Transformer \cite{zhao2021point} as the backbone and evaluate its performance on 3D point cloud classification tasks using the ModelNet40 and ShapeNet datasets, with results summarized in Tab.\ref{tab:pc}. As clearly shown in Tab.\ref{tab:pc}, our RankQ-ViT achieves better or comparable results than methods that leverage realistic point clouds for calibration on both ModelNet40 and ShapeNet datasets, showcasing its ability to generate point clouds that closely approximate the original distribution. 
PSAQ-ViT v2 performs better than PSAQ-ViT v1 on both W4/A8 and W8/A8 settings, since PSAQ-ViT v2 leverages a recycled training strategy to include more information encoded in the pretrained models.
Furthermore, our RankQ-ViT outperforms the advanced PTQ method RepQ-ViT by a clear margin of 0.49\% and 0.45\% in terms of W4/A8 and W8/A8 quantization on ModelNet40, respectively. Apart from these, our method also achieves higher results than reconstruction-based methods like BRECQ and AdaRound, which can be ascribed to the usage of the sample selection strategy. In a nutshell, these results demonstrate the superiority of our method in preserving the accuracy of point cloud models under various quantization scenarios. In contrast, calibration by Gaussian point cloud consistently exhibits the poorest performance, suggesting that the random inputs fail to preserve the geometry information of point clouds, degrading quantization efficacy. 

 \subsubsection{Results on Object Detection and Instance Segmentation}
The results in Tab.~\ref{tab:coco} demonstrate the effectiveness of the proposed RankQ-ViT across both object detection and instance segmentation tasks on the COCO dataset. Compared with the FP baseline, all quantization methods inevitably introduce performance degradation; however, RankQ-ViT consistently achieves the best trade-off between accuracy and compression. For the Cascade Mask R-CNN with Swin-S backbone, RankQ-ViT achieves 48.3 box AP and 41.7 mask AP under the W4/A8 setting, outperforming Standard quantization by +1.0/+0.8 AP and PSAQ-ViT v2 by +0.4/+0.3 AP, respectively. Notably, while Gaussian quantization suffers from severe accuracy drops (over 10 AP in both metrics), our method effectively mitigates this degradation, indicating stronger robustness to aggressive low-bit quantization. A similar trend is observed for the Mask R-CNN with Swin-S backbone, where RankQ-ViT again achieves the highest performance (45.1 box AP and 37.8 mask AP), surpassing Standard and PSAQ-ViT v2. These consistent improvements across different detection frameworks suggest that RankQ-ViT generalizes well and is not tied to a specific architecture.


\subsection{Ablation Studies}

\begin{table*}[t]
\centering
\begin{minipage}{0.33\linewidth}
\centering
\caption{Results of various loss components in RankQ-ViT. ``BA'' means the usage of block-aware mechanism. ``PSAQ'' denote the usage of losses in PSAQ-ViT v1 \cite{li2022patch}.}
\label{tab:ablation-loss}
\resizebox{\linewidth}{!}{
\begin{tabular}{cccc|cc}
\toprule
\multicolumn{4}{c}{Method} & \multicolumn{2}{c}{Prec.} \\
\midrule 
$\mathcal{L}_{rp}$ & $\mathcal{L}_{ce}$ & BA & PSAQ & 4/8 & 8/8 \\
\midrule
$\times$ &$\times$ &$\times$ &$\times$ &13.30 &18.16 \\
$\times$ &$\checkmark$ &$\times$ &$\times$ &76.57 &79.03 \\
$\times$ &$\times$ &$\times$ &$\checkmark$ &76.88 &79.12 \\
$\checkmark$ &$\times$ &$\times$ &$\times$ &77.36 &79.40 \\
$\checkmark$ &$\times$ &$\times$ &$\checkmark$ &77.62 &79.31 \\
$\times$ &$\checkmark$ &$\times$ &$\checkmark$ &77.32 &79.47 \\
$\checkmark$ &$\checkmark$ &$\times$ &$\times$ &77.68 &79.42 \\
$\checkmark$ &$\checkmark$ &$\checkmark$ &$\checkmark$ &{77.76} &{79.51} \\
$\checkmark$ &$\checkmark$ &$\checkmark$ &$\times$ &\textbf{77.86} &\textbf{79.59} \\
\bottomrule
\end{tabular}
}
\end{minipage}
\hfill
\begin{minipage}{0.33\linewidth}
\centering
\caption{Results of various sample selection and training strategies in RankQ-ViT. ``BF'' and ``NF'' mean the usage of block-wise and network-wise fine-tuning.}
\label{tab:ablation-st}
\resizebox{\linewidth}{!}{
\begin{tabular}{cccc|cc}
\toprule
\multicolumn{4}{c}{Method} & \multicolumn{2}{c}{Prec.} \\
\midrule 
BF & NF & JSD & Rank & 4/8 & 8/8 \\
\midrule
$\times$ &$\times$ &$\times$ &$\times$ &77.32 &79.47 \\
$\checkmark$ &$\times$ &$\times$ &$\times$ &77.58 &79.46 \\
$\times$ &$\checkmark$ &$\times$ &$\times$ &77.28 &79.36 \\
$\checkmark$ &$\times$ &$\checkmark$ &$\times$ &77.65 &79.46 \\
$\checkmark$ &$\times$ &$\times$ &$\checkmark$ &77.70 &79.55 \\
$\times$ &$\checkmark$ &$\checkmark$ &$\checkmark$ &77.54 &79.43 \\
$\checkmark$ &$\times$ &$\checkmark$ &$\checkmark$ &\textbf{77.86} &\textbf{79.59} \\
\bottomrule
\end{tabular}
}
\end{minipage}
\hfill
\begin{minipage}{0.33\linewidth}
\centering
\caption{Runtime (seconds) comparison on NVIDIA RTX-4090 GPU. ``Gen.'', ``FT.'', and ``Tot.'' denote the data generation, fine-tuning, and overall time, respectively.}
\label{tab:time}
\resizebox{\linewidth}{!}{
\begin{tabular}{lcccc}
\toprule
Model & Method & Gen. & FT. & Tot.\\
\midrule
\multirow{4}{*}{DeiT-T}
& PSAQ v1 & 150 & - & 150 \\
& PSAQ v2 & 202 & 216 & 418 \\
& Baseline & 28 & 216 & 244 \\
& ours & 30 & 216 & 246 \\
\midrule
\multirow{4}{*}{DeiT-S}
& PSAQ v1 & 288 & - & 288 \\
& PSAQ v2 & 414 & 474 & 888 \\
& Baseline & 46 & 474 & 520 \\
& ours & 54 & 474 & 528 \\
\bottomrule
\end{tabular}
}
\end{minipage}
\end{table*}

\subsubsection{Effectiveness of Key Components of RankQ-ViT}
To demonstrate the efficacy of the key components involved in RankQ-ViT, we conduct ablation experiments on the ImageNet dataset, utilizing DeiT-S as the network architecture. 

\begin{figure}[!bhtp]
\centering
\centering
\includegraphics[width=0.98\linewidth]{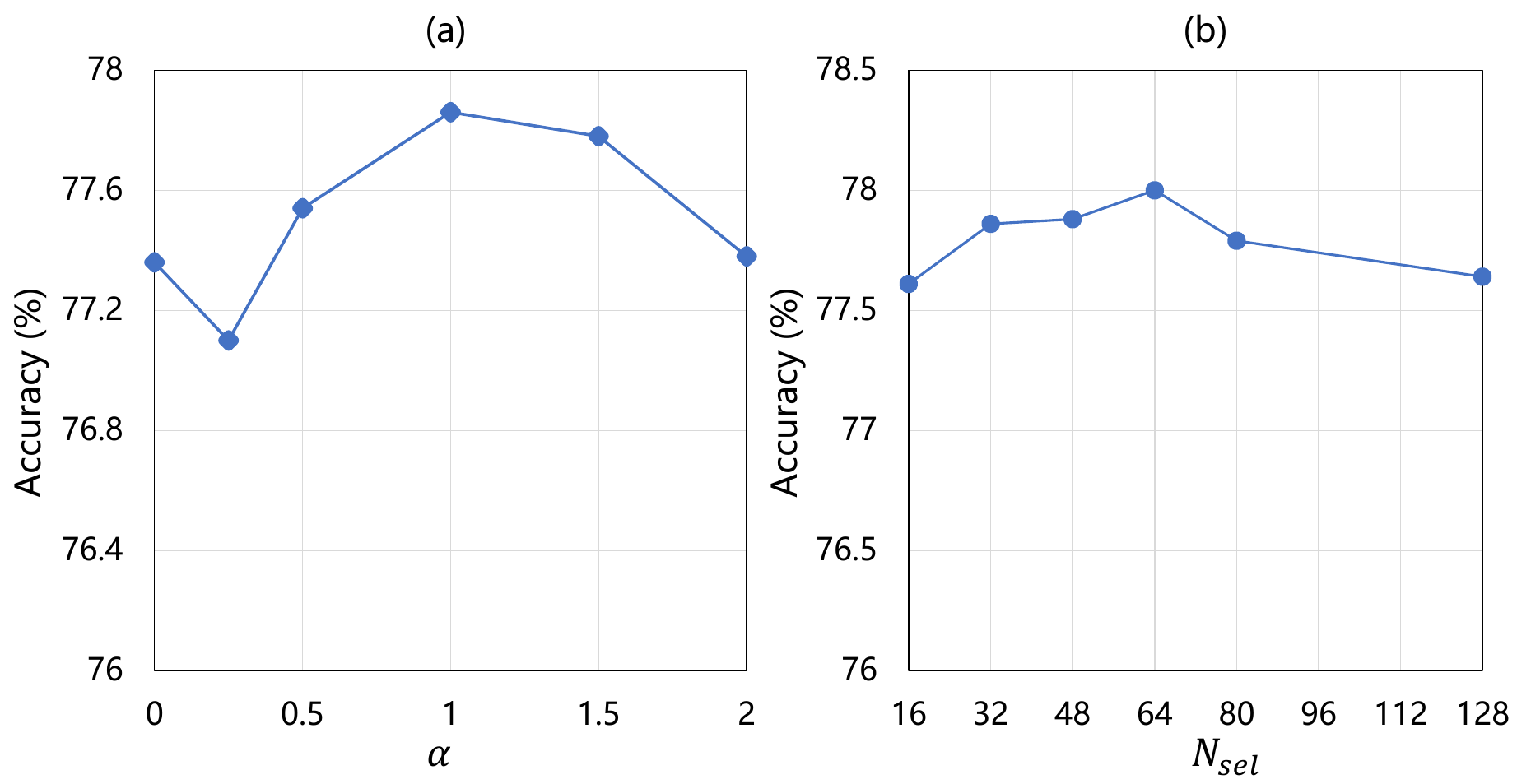}
\caption{ Hyperparameter sensitivity analysis on W4/A8 quantization of DeiT-S. (a) Effect of different $\alpha$ values. (b) Effect of different $N_{sel}$ values.}
\label{fig:ab-alpha-nsel}
\hfill
\end{figure}
First, we evaluate the effectiveness of the loss functions used in our method, as shown in Tab.~\ref {tab:ablation-loss}. The model exhibits the worst performance when all loss variants are excluded, highlighting the importance of each component in achieving optimal results. Notably, the PSAQ loss performs slightly better than the sole usage of $\mathcal{L}_{ce}$ due to the entropy maximization operation. Our $\mathcal{L}_{rp}$ outperforms the PSAQ by 0.48\% and 0.28\% in terms of W4/A8 and W8/A8, showing the superiority of our rank promotion strategy in preserving the nuances of patch relationships. Integrating $\mathcal{L}_{rp}$ with PSAQ yields superior performance compared to the baseline PSAQ, as $\mathcal{L}_{rp}$ facilitates the extraction of more discriminative features. We also note that combining $\mathcal{L}_{rp}$ with PSAQ does not yield better improvement over rank alone, but rank alone already outperforms PSAQ alone, which suggests that rank is a more direct and effective optimization target.


Next, we validate the sample selection and training strategies used in our method, as shown in Tab.~\ref{tab:ablation-st}. Specifically, the NF and BF aim to match the features between the original network and its low-bit counterpart in a network-wise and block-wise respectively. When replacing BF with NF, the test accuracy decreases by 0.32\% and 0.16\% for W4/A8 and W8/A8, respectively. This can be attributed to the overfitting that occurs during FT \cite{librecq}. Moreover, using rank in sample selection improves the accuracy, indicating that samples with higher ranks tend to have better quality. By incorporating JSD and rank together, our method achieves the best results for both W4/A8 and W8/A8 quantization. This suggests that the quality of generated images can be effectively assessed through intermediate feature distribution and semantic consistency.

\subsubsection{Sensitivity Analysis of Hyper-parameters} We first investigate the influence of varied $\alpha$ in Eq.~\ref{eq:total}. From Fig.~\ref{fig:ab-alpha-nsel} (a), it can be easily observed that models receive the best performance when $\alpha=1$. When $\alpha=0$, which means the absence of $\mathcal{L}_{ce}$, the models have the worst performance since the loss of semantic supervision. In addition, when shifting the $\alpha$ from 1 to 2, there is a slight accuracy loss, indicating the robustness of $\alpha$ during training.
We also evaluate the impact of different $N_{sel}$, which is shown in Fig.~\ref{fig:ab-alpha-nsel} (b). We observe that selecting too few samples for calibration leads to suboptimal results, due to the overfitting during block-wise reconstruction. However, exclusively using all the generated samples also shows limited results, as it overlooks the quality of the samples. Therefore, we use a moderate portion of generated samples for fine-tuning stage.

\subsubsection{Visualization Analysis of Generated samples} 
We visualize the samples generated by DeiT-T in Fig.~\ref{fig:syn-image}. The images produced by PSAQ-ViT exhibit lower fidelity than those generated by RankQ-ViT. Moreover, images generated by  PSAQ-ViT contain less semantic information than our RankQ-ViT. For instance, a clear outline of the object labeled ``minivan'' is discernible in the generated image by RankQ-ViT. These results suggest that our rank promotion strategy effectively preserves feature independence, thereby leading to improved image generation quality.

\subsubsection{Effectiveness of Cross-Architecture Evaluation} 
\begin{figure}[] 
	\centering 
	\includegraphics[width=0.5\textwidth]{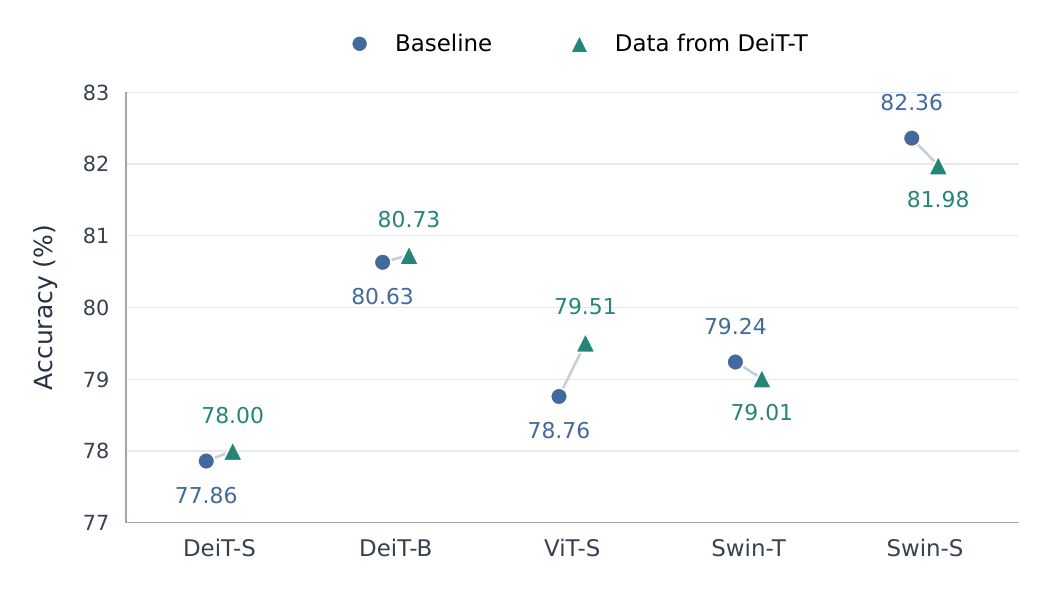} 
	\caption{Results of data generated by DeiT-T for other model quantization. W4/A8 quantization is used here for comparison. The results clearly indicate that our RankQ-ViT has superior cross-architecture generalization ability.} \label{fig:cross-arch}
\end{figure}

To assess the generalization capability of the generated images, we further conduct cross-architecture calibration experiments, using images generated by DeiT-T as source data to calibrate other network variants. Specifically, we fix the synthetic samples produced by DeiT-T and directly apply them to fine-tune the quantized versions of DeiT-S, DeiT-B, and ViT-S without any further data generation. As shown in Fig.~\ref{fig:cross-arch}, the results demonstrate that data generated by DeiT-T consistently improves the performance of DeiT-S, DeiT-B, and ViT-S. This finding validates the robustness and generalization of the proposed method across different architectures, suggesting that our rank-promoted synthetic data captures transferable feature distributions rather than model-specific artifacts.


\subsubsection{Efficiency Analysis of RankQ-ViT}
In addition to the test performance, the quantization efficiency is also critical in realistic applications. To ensure a fair comparison, we fix the batch size at 32 and the total iterations at 2,000 for all variants. The results are detailed in Tab.~\ref{tab:time}. The PSAQ-ViT v2 struggles in efficiency due to the two-stream learning in data generation. The data generation process in RankQ-ViT is approximately 5 times faster than PSAQ-ViT v1, since it eliminates the need for complex entropy computation. 
Overall, our RankQ-ViT strikes a good balance between efficiency and performance, offering great potential for practical usage.

\subsubsection{Deployment of Quantized Models}   
To evaluate the deployment efficiency of quantized ViT models on general-purpose computation platforms (i.e., CPU/GPU), we benchmark ViT-B and DeiT-B using W8/A8 quantization during inference. 
For CPU deployment, we use 8 CPU inference threads for implementation. 
For GPU deployment, we use TensorRT-based inference with FP32 and INT8 engines, and report throughput and peak GPU memory usage measured during deployment. We use 20 warm-up iterations followed by 100 timed inference runs, and throughput was computed from the median latency.

\begin{figure}[!t] 
	\centering 
	\includegraphics[width=0.5\textwidth]{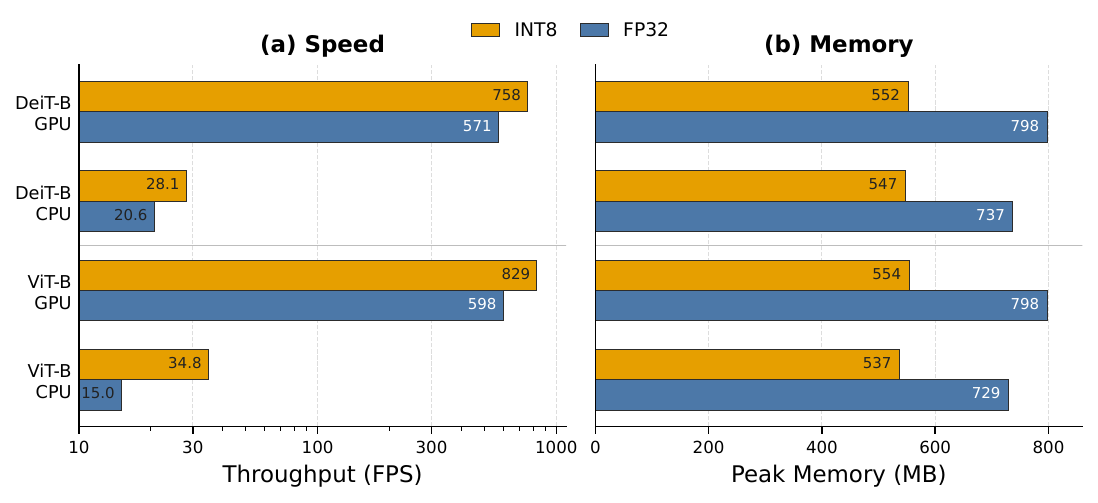} 
	\caption{Speed (FPS) and peak memory (MB) usage during inference on Intel 12th i9-12900K CPU and NVIDIA RTX-4090 GPU. 
    (a) FPS comparison between FP32 and INT8 models. 
    (b) The Peak GPU memory during inference. These results show that quantization significantly reduces the inference cost.} \label{fig:infer}
\end{figure}

On CPU, INT8 quantization improves both throughput and memory efficiency. For ViT-B, throughput increases from 15.0 FPS to 34.8 FPS, giving a 2.31× speedup, while peak memory decreases from 729 MB to 537 MB. For DeiT-B, throughput increases from 20.6 FPS to 28.1 FPS, corresponding to a 1.36× speedup, while peak memory decreases from 737 MB to 547 MB. 
The GPU TensorRT results show a similar but much larger absolute throughput advantage. 
For ViT-B, INT8 TensorRT inference improves throughput from 598 FPS to 829 FPS and reduces peak memory from 798 MB to 554 MB. 
For DeiT-B, throughput increases from 571 FPS to 758 FPS, while peak memory decreases from 798 MB to 552 MB. 
These results demonstrate that INT8 quantization consistently reduces memory consumption and improves inference throughput across both CPU and GPU deployment backends.

\section{Conclusion and Future Work}
This paper proposes RankQ-ViT, an efficient data-free quantization method tailored for Vision Transformers (ViTs). RankQ-ViT employs a two-step training framework, which first generates training samples and then fine-tunes the model using these samples. Specifically, we use a novel block-aware rank promotion mechanism, which encourages the generated images to closely mimic the rank distribution of realistic ones in a block-aware fashion. In this way, high-quality samples can be generated efficiently. After that, RankQ-ViT selects high-quality samples with the highest semantic consistency and fidelity, thereby enhancing the quantization performance effectively. Extensive experiments on four datasets (including ImageNet, COCO, ModelNet40, and ShapeNet) for image classification, object detection, instance segmentation, and 3D point cloud classification tasks demonstrate the effectiveness of the proposed RankQ-ViT. With the W4/A8 quantization setting, the proposed RankQ-ViT achieves 82.44\% and 77.86\% test accuracy on ViT-B and DeiT-S, which significantly outperforms the state-of-the-art PTQ method RepQ-ViT by 6.15\% and 2.67\%. 
As future work, we plan to explore data-free quantization methods for large-scale foundation models (e.g., SAM), to facilitate the deployment of these models.

\section*{Acknowledgments}
This work was partially supported by the Guangdong S\&T Programme (No. 2025B0101130003), the National Natural Science Foundation of China (No. 62372491), the Guangdong Basic and Applied Basic Research Foundation (2023B1515120087), the Aeronautical Science Fund of China (2025L0150M1001), and the Science and Technology Planning Project of Key Laboratory of Advanced IntelliSense Technology, Guangdong Science and Technology Department (2023B1212060024).

\bibliographystyle{IEEEtran}
\bibliography{ref}
\vspace{5pt}


\appendix 
\subsection{The proof of Proposition~\ref{them:rank}} \label{prof:rank}
\begin{proof}
For brevity, we use $r=2$ in the proof process. Consider feature vectors $x_i\in\mathbb{R}^{D}$, which is defined as
\begin{equation}
x_i=
(\cos\theta_i,\sin\theta_i,0,\dots,0),
\quad
\theta_i=\frac{2\pi i}{N},
\quad i=1,\dots,N,
\end{equation}
where only the first two coordinates are nonzero, and represented in a triangle function style.
Then, let
\begin{equation}
X=
[x_1^\top,\dots,x_N^\top]^\top
\in\mathbb{R}^{N\times D},
\end{equation}
and define the similarity matrix
\begin{equation}
S_N=XX^\top.
\end{equation}
Since all feature vectors lie in a two-dimensional subspace of $\mathbb{R}^{D}$, we have
\begin{equation}
\operatorname{rank}(X)\le 2.
\end{equation}
Therefore,
\begin{equation}
\operatorname{rank}(S_N)
=
\operatorname{rank}(XX^\top)
\le
\operatorname{rank}(X)
\le 2.
\end{equation}
For any pair $(i,j)$, the similarity entry satisfies
\begin{equation}
S_{ij}
=
x_i^\top x_j
=
\cos(\theta_i-\theta_j).
\end{equation}
As $N\rightarrow\infty$, the angle differences $\theta_i-\theta_j$ become dense in $[0,2\pi]$, implying that the similarity values $\{S_{ij}\}$ become dense in $[-1,1]$.
Therefore, although $S_N$ remains low-rank, its entries are highly dispersed across the entire similarity range, leading to high entropy. 




\end{proof}

\subsection{The proof of Proposition~\ref{them:1}} \label{prof:1}
\begin{proof}
We first define the Gram matrix as
\begin{equation}
G = XX^\top .
\end{equation}
Since each row is normalized (feature is normalized), the diagonal entries of $G$ satisfy
\begin{equation}
G_{ii} = x_i^\top x_i = 1.
\end{equation}
For every $i\neq j$, according to the assumption, it gives:
\begin{equation}
|G_{ij}| = |x_i^\top x_j| \le \epsilon .
\end{equation}
Thus the Frobenius norm of $G$ satisfies
\begin{align}
\|G\|_F^2
&= \sum_{i,j} G_{ij}^2 \\
&= \sum_i G_{ii}^2 + \sum_{i\neq j} G_{ij}^2 \\
&\le N + N(N-1)\epsilon^2 .
\end{align}
Let $\lambda_1,\dots,\lambda_r$ denote the non-zero eigenvalues of $G$, where
\begin{equation}
r=\mathrm{rank}(G)=\mathrm{rank}(X).
\end{equation}
We have
\begin{equation}
\mathrm{tr}(G) = \sum_{i=1}^{r} \lambda_i = N, \|G\|_F^2 = \sum_{i=1}^{r} \lambda_i^2  .
\end{equation}
And according to the Cauchy–Schwarz inequality, we have
\begin{equation}
\left(\sum_{i=1}^{r} \lambda_i\right)^2 
\le r \sum_{i=1}^{r} \lambda_i^2 .
\end{equation}
Then, substituting the above identities yields
\begin{equation}
N^2 \le r \|G\|_F^2 .
\end{equation}
Therefore
\begin{equation}
r \ge \frac{N^2}{\|G\|_F^2}.
\end{equation}
Finally, combining with the bound on $\|G\|_F^2$, we have
\begin{equation}
r \ge \frac{N^2}{N+N(N-1)\epsilon^2}
= \frac{N}{1+(N-1)\epsilon^2}.
\end{equation}
Since $r=\mathrm{rank}(X)=\mathrm{rank}(G)$, the theorem follows. 
\end{proof}

\end{document}